\documentclass[runningheads]{llncs}

\usepackage{graphicx}
\usepackage{amsmath,amssymb} 
\usepackage{algorithm}
\usepackage{algpseudocode}
\usepackage[hidelinks]{hyperref} 

\usepackage{listings}
\title{Measuring Uncertainty in Transformer Circuits with Effective Information Consistency}
\titlerunning{Effective Information Consistency for Transformer Circuits}

\author{Anatoly A. Krasnovsky\inst{1,2}}
\authorrunning{A.~A. Krasnovsky}
\institute{Department of Computer Science and Engineering, Innopolis University, Russia \\
\and
MB3R Lab, Innopolis, 420500, Russia}

\spnewtheorem{assumption}{Assumption}{\bfseries}{\itshape}

\begin{document}
\maketitle

\begin{abstract}
Mechanistic interpretability has identified functional subgraphs within large language models (LLMs), known as Transformer Circuits (TCs), that appear to implement specific algorithms. Yet we lack a formal, single-pass way to quantify when an \emph{active} circuit is behaving coherently and thus likely trustworthy. Building on prior systems-theoretic proposals, we specialize a sheaf/cohomology and causal emergence perspective to TCs and introduce the \emph{Effective-Information Consistency Score (EICS)}. EICS combines (i) a \emph{normalized sheaf inconsistency} computed from local Jacobians and activations, with (ii) a \emph{Gaussian EI proxy} for circuit-level causal emergence derived from the same forward state. The construction is white-box, single-pass, and makes units explicit so that the score is dimensionless. We further provide practical guidance on score interpretation, computational overhead (with fast and exact modes), and a toy sanity-check analysis. Empirical validation on LLM tasks is deferred.
\end{abstract}

\section{Introduction}
A central ambition in mechanistic interpretability is to reverse-engineer LLMs into meso-scale algorithmic components (``transformer circuits'') \cite{olsson2022,attributiongraphs2025}. These subgraphs (attention heads, MLPs, and their pathways) have been linked to tasks such as copy/induction \cite{olsson2022} and factual recall \cite{yao2024}. Once a circuit is identified, a natural question is: \emph{is it functioning coherently on this input?} The same model may answer a factual prompt correctly or hallucinate; we hypothesize that the active circuit's \emph{causal integrity} differs between these regimes.

We build on sheaf-theoretic and causal-emergence ideas \cite{krasnovsky2025,hansen2019,robinson2014tsp,rosas2020,tononi2003,oizumi2014}, instantiating them for transformer circuits. Our contribution is an \emph{operational} metric, EICS, that can be computed from a single forward pass. Conceptually, uncertainty is treated as a failure of causal integrity: high emergence with low internal inconsistency suggests confidence; the opposite suggests risk. Unlike black-box UQ \cite{angelopoulos2021,guo2017,lakshminarayanan2017,yang2023}, EICS attributes uncertainty to specific \emph{mechanisms}.

\section{Related Work}
\textbf{Mechanistic interpretability.} Induction heads and in-context learning behaviors were documented by Olsson et al.\ \cite{olsson2022}. Recent ``attribution graphs'' tools (a.k.a.\ circuit tracing) provide subgraph discovery and validation pipelines \cite{attributiongraphs2025}. Knowledge-related circuits are actively studied \cite{yao2024}, and the relation between localization and editing has been scrutinized \cite{hase2023}.

\textbf{Uncertainty quantification (UQ).} Black-box UQ includes distribution-free conformal prediction \cite{angelopoulos2021}, calibration post-processing \cite{guo2017}, deep ensembles \cite{lakshminarayanan2017}, and Bayesian/approximate Bayesian methods for LLM finetuning such as Laplace-LoRA \cite{yang2023}. Our aim is complementary: a \emph{white-box} signal grounded in the internal circuit.

\textbf{Sheaves and causal emergence.} Cellular sheaves provide a language for stitching local linear maps into globally consistent states; disagreement is measured via coboundaries and Hodge Laplacians \cite{robinson2014tsp,hansen2019}. Causal emergence quantifies when macroscale descriptions carry more effective information than parts \cite{rosas2020,tononi2003,oizumi2014}. We adapt these ingredients to transformer circuits with explicit, computable proxies.

\section{Preliminaries}

\subsection{Transformer circuits}
Let the transformer be a directed acyclic graph (DAG) $G=(V,E)$ with vertices as heads/MLPs and edges for information flow. A TC is a subgraph $G_M\subseteq G$ hypothesized to implement a task. For an input $x$, each $v\in V_M$ has activation $a_v\in\mathbb{R}^{d_v}$. We will write $e=(u\!\to\!v)$ for an oriented edge and use local linearizations at the observed activations.

\subsection{Cellular sheaves on graphs and a computable inconsistency}
\label{sec:sheaf}
We place a cellular sheaf $\mathcal{F}$ on the \emph{underlying undirected} version of $G_M$ (so that 1-cochains live on edges regardless of the DAG's direction), with stalks $\mathcal{F}(v)=\mathbb{R}^{d_v}$ and restriction maps on oriented edges given by the Jacobians evaluated at the current state:
\[
\rho_{e}: \mathcal{F}(u)\to \mathcal{F}(v),\qquad
\rho_{u\to v} \;:=\; J_{u\to v} \;\equiv\; \Big(\frac{\partial f_v}{\partial a_u}\Big)_{a}.
\]
For a 0-cochain (node assignment) $s=\{s_v\}$, the sheaf coboundary $\delta^0:\!C^0\!\to\!C^1$ acts by
\begin{equation}
(\delta^0 s)_{(u\to v)} \;=\; \rho_{u\to v}\, s_u \;-\; s_v.
\end{equation}
On a graph (no 2-cells), $\delta^1=0$, hence $H^1 \cong C^1/\mathrm{im}\,\delta^0$. Rather than taking a non-canonical ``norm of a quotient,'' we use a \emph{normalized inconsistency energy} from the observed activations $a=\{a_v\}$:
\begin{equation}
\label{eq:Csh}
C_{\mathrm{sh}}(G_M,a)\;=\;
\frac{\left(\sum_{(u\to v)\in E_M}\|\rho_{u\to v}a_u - a_v\|_2^2\right)^{1/2}}
{\varepsilon + \left(\sum_{(u\to v)\in E_M} \|a_u\|_2^2 + \|a_v\|_2^2\right)^{1/2}},
\end{equation}
with small $\varepsilon>0$ for numerical stability. This is dimensionless and equals $0$ iff $a$ is a (noisily) consistent global section. One can optionally replace $a$ by the least-squares projection $\hat s=\arg\min_s \sum_e\|\rho_e s_u-s_v\|^2$; both are single-pass (VJP/JVP-based) computations.

\paragraph{Node-seeded JVP computation (efficiency).}
To evaluate \eqref{eq:Csh} efficiently, we use a \emph{node-seeded} JVP scheme: for each source node $u\in V_M$ we perform a single JVP with seed $a_u$ to evaluate all outgoing residual terms $\rho_{u\to v} a_u$ in one pass. This reduces the complexity from per-edge JVPs to $O(|V_M|)$ JVPs (typically $10$--$20$ for medium circuits), while preserving the exact definition in \eqref{eq:Csh}.

\subsection{A single-pass, Gaussian EI proxy}
True effective information (EI) is intervention-defined. To obtain a single-pass proxy, we assume small, isotropic local interventions at the current state and approximate each map by its Jacobian. For a linear map $y=Jx+\xi$ with unit-variance isotropic $x$ and small additive noise $\xi$, the mutual information (in nats) is proportional to $\tfrac12 \log\det(I + \alpha J^\top J)$ for scale $\alpha>0$. We therefore define
\begin{align}
\label{eq:EIproxy}
\mathrm{EI}_G(J) \;:=\; \tfrac12 \log\det\!\big(I + \alpha J^\top J\big),\qquad
\Delta \mathrm{EI}_G(G_M) \;:=\; \mathrm{EI}_G(J_{M}) \;-\; \sum_{v\in V_M}\mathrm{EI}_G(J_v),
\end{align}
where $J_M$ is the macro-Jacobian from the circuit's inputs to its outputs (obtained by linearizing the composed subgraph). We use the positive part $\Delta \mathrm{EI}_G^+=\max(0,\Delta \mathrm{EI}_G)$, and an optional normalization $\widetilde{\Delta \mathrm{EI}}_G := \Delta \mathrm{EI}_G^+ / \big(\varepsilon + \mathrm{EI}_G(J_M)\big)$ to keep scores in $[0,1)$.

\subsection{Gaussian EI Proxy — Derivation and Implementation Notes}
\label{sec:gaussian-ei-derivation}
\paragraph{Setup.} Consider a local, linearized description of a circuit around the observed forward state. Let $x\in\mathbb{R}^n$ denote a small stochastic intervention at the circuit inputs, and let $y\in\mathbb{R}^m$ be the circuit outputs. We approximate
\begin{equation}
y \;=\; J\,x + \xi,\qquad x\sim\mathcal{N}(0,\sigma_x^2 I_n),\ \ \xi\sim\mathcal{N}(0,\sigma_\xi^2 I_m).
\end{equation}
\paragraph{Mutual information.} For a linear Gaussian channel with independent Gaussian input and noise,
\begin{equation}
I(x;y) \;=\; \tfrac{1}{2}\log\det\!\Big(I_m + \tfrac{\sigma_x^2}{\sigma_\xi^2} J J^\top\Big)
\;=\; \tfrac{1}{2}\log\det\!\Big(I_n + \tfrac{\sigma_x^2}{\sigma_\xi^2} J^\top J\Big).
\end{equation}
\paragraph{Proxy definition.} We use the Gaussian EI proxy from Eq.~\eqref{eq:EIproxy}; for a circuit $G_M$
the emergence and its normalized positive part are, as in §3.3,
\begin{equation}
\widetilde{\Delta \mathrm{EI}}_G \;=\; \frac{\max(0,\Delta \mathrm{EI}_G)}{\varepsilon + \mathrm{EI}_G(J_M)}.
\end{equation}
\paragraph{Invariance, sensitivity, and small-$\alpha$ approximation.}
$\mathrm{EI}_G(J)$ depends only on the singular values of $J$. Its sensitivity to $\alpha$ is
\[
\frac{\partial}{\partial \alpha}\Big(\tfrac12 \log\det(I+\alpha J^\top J)\Big)
= \tfrac12\,\mathrm{tr}\!\big[(I+\alpha J^\top J)^{-1} J^\top J \big],
\]
and for $\alpha\,\sigma_{\max}^2\ll 1$,
\[
\tfrac12\log\det(I+\alpha J^\top J) \;\approx\; \tfrac{\alpha}{2}\,\|J\|_F^2.
\]
\paragraph{Computation.} Use Cholesky/eigendecomposition for small $n$; for larger $n$, use Hutch++/Lanczos log-det estimators with only JVP/VJP products. Residual connections can be handled by building the linearized block operator or evaluating log-det via Krylov methods.

\section{Method: EICS}
Given $G_M$, activations $a$, and edge Jacobians $\{\rho_{u\to v}\}$ from a single forward pass, we define
\begin{equation}
\label{eq:eics}
\mathrm{EICS}(G_M;a)
\;=\;
\frac{\widetilde{\Delta \mathrm{EI}}_G(G_M)}{1 + C_{\mathrm{sh}}(G_M,a)}.
\end{equation}
High EICS indicates (i) strong macro-level information integration relative to parts and (ii) low internal disagreement across edges.

\paragraph{Why this fixes earlier issues.} (1) We never take a norm of a quotient space $H^1$; we measure \emph{disagreement energy} \eqref{eq:Csh} directly and dimensionlessly. (2) The EI term is a clearly stated Gaussian-logdet proxy; units are nats and become dimensionless via normalization. (3) DAGs have no directed cycles, but sheaf inconsistency remains meaningful on the undirected 1-skeleton.

\subsubsection*{Practical use and interpretation.}
By construction $C_{\mathrm{sh}}\!\ge0$ and $\widetilde{\Delta \mathrm{EI}}_G\in[0,1)$, hence $\mathrm{EICS}\in[0,1)$. Use a development split to select a threshold $\tau$ (AUROC/F1) and also report the components $1/(1+C_{\mathrm{sh}})$ and $\widetilde{\Delta \mathrm{EI}}_G$ to diagnose drivers. For $\alpha$, either set $\alpha=1$ (SNR prior) or choose $\alpha$ to keep $\tfrac12\log\det(I+\alpha J_M^\top J_M)$ in a target IQR to avoid saturation.

\section{Theoretical properties}

\begin{assumption}[Local linearity and boundedness]
\label{assump:lin}
Along $G_M$, the maps are locally linear with Jacobians $\{\rho_e\}$ Lipschitz in a neighborhood of $a$, and operator norms are bounded. The sheaf Hodge Laplacian $L=\delta^{0\dagger}\delta^0$ (with the inner product induced by edge weights) has spectral gap $\lambda_2(L)>0$ \textup{\cite{hansen2019}}.
\end{assumption}

\begin{proposition}[Single-pass computability]
Under Assumption~\ref{assump:lin}, both $C_{\mathrm{sh}}(G_M,a)$ and $\widetilde{\Delta \mathrm{EI}}_G(G_M)$ are deterministic functions of a single forward pass and its Jacobian-vector products. Consequently, $\mathrm{EICS}$ is $O(1)$ in the number of forward passes.
\end{proposition}

\noindent\textit{Sketch.} $\delta^0$ is built from $\{\rho_e\}$ evaluated at $a$. Both the residual \eqref{eq:Csh} and log-det terms \eqref{eq:EIproxy} are functions of these objects. No Monte Carlo over inputs is required.

\begin{proposition}[Stability to small off-circuit perturbations (bound)]
\label{prop:stability}
Let $y$ denote circuit outputs. Consider an additive off-circuit perturbation $\eta$ that couples into $G_M$ with gain at most $\gamma$ in operator norm. Under Assumption~\ref{assump:lin},
\[
\|\hat s - s^\star\| \;\le\; \frac{\gamma}{\lambda_2(L)}\;\|\eta\|,
\qquad
\| \Delta y \| \le \kappa \| \hat s - s^\star \|
\]
for a local Lipschitz constant $\kappa$. In particular, small $C_{\mathrm{sh}}$ (implying a large effective $\lambda_2(L)$) yields tighter bounds.
\end{proposition}

\noindent\textit{On the role of $\lambda_2(L)$.}
The bound scales with $1/\lambda_2(L)$ (Fiedler value). In practice $\lambda_2(L)$ depends on (i) connectivity and (ii) edge weights induced by local Jacobians. We recommend: (a) \emph{report} $\lambda_2(L)$ per circuit; (b) \emph{normalize edge weights} by operator norms of $\rho_{u\to v}$; and (c) optionally regularize $L\leftarrow L+\beta I$ when empirical $\lambda_2(L)$ is near-zero—tightening the practical bound without changing $C_{\mathrm{sh}}$ or EICS.

\section{Algorithm}
\begin{algorithm}
\caption{Single-pass EICS for a transformer circuit}
\label{alg:eics}
\begin{algorithmic}[1]
\State \textbf{Input:} Model $\mathcal{M}$, input $x$, circuit $G_M=(V_M,E_M)$, scale $\alpha>0$
\State \textbf{Output:} $\mathrm{EICS}(G_M;a)$
\State \textbf{Forward \& activations:} Run $\mathcal{M}(x)$ and record $\{a_v\}_{v\in V_M}$.
\State \textbf{Edge Jacobians:} For each $(u\to v)\in E_M$, compute $\rho_{u\to v} = \big(\partial f_v / \partial a_u\big)_a$ via VJP/JVP.
\State \textbf{Sheaf inconsistency:} Compute $C_{\mathrm{sh}}(G_M,a)$ using \eqref{eq:Csh}. \emph{Implementation:} Use \emph{node-seeded JVPs} (one JVP per source node $u$) to evaluate all $\rho_{u\to v}a_u$ for outgoing edges.
\State \textbf{Gaussian EI proxy:} Build macro-Jacobian $J_M$ and node Jacobians $\{J_v\}$. Compute $\widetilde{\Delta \mathrm{EI}}_G$ via \eqref{eq:EIproxy}.
\Statex \quad \textbf{Fast mode (ranking):} small-$\alpha$ approximation; Hutch++/Lanczos with 4--8 probes per $J_v$ and 8--12 for $J_M$.
\Statex \quad \textbf{Exact mode (small blocks):} compute $\log\det(I+\alpha J^\top J)$ via Cholesky/SVD.
\State \textbf{Score:} Return $\mathrm{EICS}=\widetilde{\Delta \mathrm{EI}}_G/(1+C_{\mathrm{sh}})$.
\end{algorithmic}
\end{algorithm}

\subsection*{Computational overhead \& scaling}
\textbf{Order-of-magnitude.} With node-seeded JVPs, computing $C_{\mathrm{sh}}$ requires $O(|V_M|)$ JVPs (about $10$--$20$ for medium circuits). In \emph{fast} EI mode (small-$\alpha$ Frobenius approximation + Hutchinson-style probes), total autograd work is typically $\sim 50$--$200$ JVP/VJP products over the restricted subgraph, or about $\sim 2$--$6$ forward-equivalents. In \emph{exact} EI mode (larger $\alpha$ or explicit factorizations), expect $\sim 5$--$15$ forward-equivalents. There is no input-level Monte Carlo; all quantities are deterministic functions of one forward state.

\textbf{Scaling knobs.} (a) Batch JVP seeds across nodes; (b) restrict EI to top-$k$ singular directions via Lanczos; (c) cache intermediate linears along the TC; (d) prefer small-$\alpha$ for ranking tasks.

\section{Proposed Validation}
We outline an evaluation protocol for a factual QA task using an attribution-graph workflow \cite{attributiongraphs2025} to identify a fact-retrieval circuit $G_{\mathrm{fact}}$. Two sets are proposed: (A) questions with verifiable answers; (B) adversarial or hallucination-inducing prompts \cite{huang2023}. The expectation is higher $\mathrm{EICS}$ for set~A (low $C_{\mathrm{sh}}$, positive $\widetilde{\Delta \mathrm{EI}}_G$) and depressed scores for set~B. Comparators include log-prob/entropy, deep-ensemble variance \cite{lakshminarayanan2017}, and conformal set sizes \cite{angelopoulos2021}.

\subsection{Baselines and ablations}
\textbf{(B1) Edge Activation Correlation (EAC).} Average Pearson correlation between $a_u$ and $a_v$ across $(u\!\to\!v)\in E_M$ (dimension-wise). \\
\textbf{(B2) Edge Alignment Residual (EAR).} $\frac{1}{|E_M|}\sum_{(u\to v)} \| \hat{\rho}_{u\to v} a_u - a_v\|_2$ with per-edge least-squares $\hat{\rho}_{u\to v}$ (no sheaf coupling). \\
\textbf{Ablations.} (A1) $1/(1+C_{\mathrm{sh}})$ only; (A2) $\widetilde{\Delta \mathrm{EI}}_G$ only.

\subsection{Toy sanity-check (analytical \& simulation protocol)}
\textbf{Setup.} A 6-node feedforward TC with linear blocks and additive Gaussian edge noise; vary noise on a subset of edges. Compute $C_{\mathrm{sh}}$, $\widetilde{\Delta \mathrm{EI}}_G$ (fast/exact), EICS, and baselines across seeds.

\begin{figure}[t]
  \centering
  \includegraphics[width=.9\linewidth]{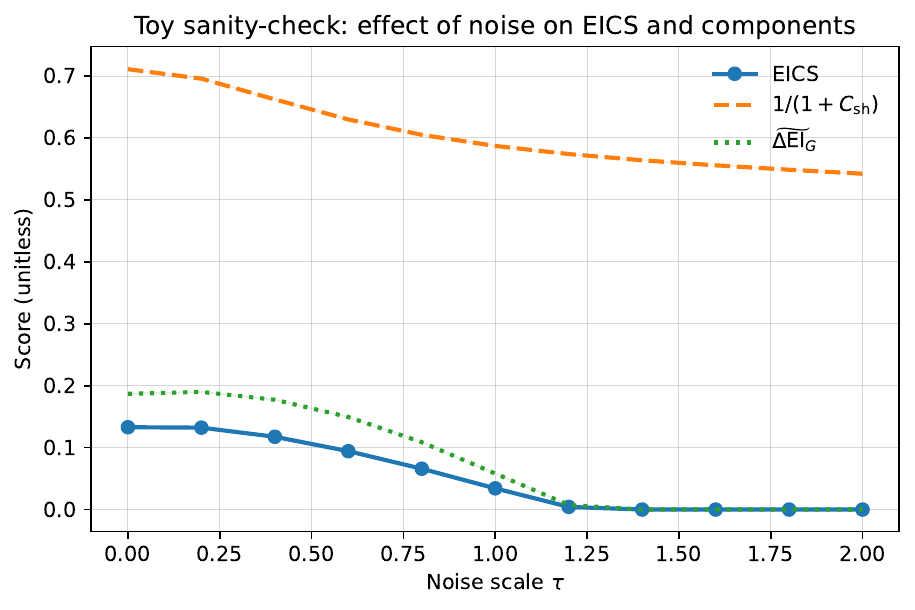}
  \caption{Toy sanity-check on a 6-node circuit with two parallel branches. As node-noise $\tau$ increases, the sheaf inconsistency $C_{\mathrm{sh}}$ rises (so $1/(1+C_{\mathrm{sh}})$ falls). We also reduce cross-branch alignment with $\tau$ (edge decoherence), causing the emergence proxy $\widetilde{\Delta \mathrm{EI}}_G$ and the overall EICS to decrease. Curves show means over seeds (no error bands for clarity). Definitions follow Eqs.~\eqref{eq:Csh}, \eqref{eq:EIproxy}, and \eqref{eq:eics}.}
  \label{fig:toy}
\end{figure}

\textbf{Analytical sanity (small-$\alpha$).} With $\alpha=\sigma_x^2/\sigma_\xi^2$, increasing additive noise reduces EI terms; edge noise increases residuals in \eqref{eq:Csh}. Hence EICS decreases with noise, matching intuition. See Fig.~\ref{fig:toy} for a synthetic sanity-check illustrating these trends.

\section{Discussion \& limitations}
\textbf{Circuit dependency.} EICS is only as meaningful as the specified $G_M$. \textbf{Linearization.} The Jacobian-based sheaf and EI proxy assume local linearity. \textbf{Cost.} Node-seeded JVPs, fast log-det, and top-$k$ directions mitigate overhead. \textbf{Role with black-box UQ.} EICS provides \emph{mechanistic} evidence complementing calibration/conformal tools.

\section{Conclusion}
We instantiate a sheaf-theoretic and causal-emergence perspective on transformer circuits as a practical, single-pass score (EICS). By replacing ill-posed cohomology norms with a normalized disagreement energy and adopting a Gaussian log-det EI proxy, EICS is both computable and dimensionless. We add practical guidance, computational cost analysis with fast/exact modes, and a toy sanity-check; full empirical validation follows the proposed protocol.

\appendix
\section{Toy Sanity-Check Code for Fig.~\ref{fig:toy}}
\noindent
The script below reproduces Fig.~\ref{fig:toy}. It implements the two-branch toy circuit, computes
$C_{\mathrm{sh}}$ (Eq.~\eqref{eq:Csh}), the normalized emergence proxy $\widetilde{\Delta \mathrm{EI}}_G$
(Eqs.~\eqref{eq:EIproxy}), and $\mathrm{EICS}$ (Eq.~\eqref{eq:eics}) as functions of the noise scale~$\tau$.

\begin{lstlisting}
# Minimal toy-plot script for Fig. "Toy sanity-check"
# Requires: numpy, matplotlib. Saves: fig_toy_noise_curve.pdf
import numpy as np, matplotlib.pyplot as plt
EPS, D, N_SEEDS = 1e-8, 32, 100
TAUS = np.linspace(0.0, 2.0, 11)
alpha, align = 1.0, 0.9  # fixed SNR; high initial branch alignment

def rand_matrix(d, scale=0.8, rng=None):
    rng = np.random.default_rng() if rng is None else rng
    return scale * rng.normal(size=(d, d)) / np.sqrt(d)

def ei_proxy(J, alpha):  # 0.5 * sum log(1 + alpha * sigma^2)
    s = np.linalg.svd(J, compute_uv=False)
    return 0.5 * np.sum(np.log1p(alpha * (s**2)))

def build_branch_mats(D=32, align=0.9, rng=None):
    rng = np.random.default_rng(123) if rng is None else rng
    U = rand_matrix(D, 0.8, rng); A = rand_matrix(D, 0.9, rng); W = rand_matrix(D, 0.9, rng)
    W13 = U;  W23 = (1-align)*rand_matrix(D,0.8,rng) + align*U
    W34 = A;  W35 = (1-align)*rand_matrix(D,0.9,rng) + align*A
    W46 = W;  W56 = (1-align)*rand_matrix(D,0.9,rng) + align*W
    return W13, W23, W34, W35, W46, W56

def metrics_at_tau(tau, rng):
    W13,W23,W34,W35,W46,W56 = build_branch_mats(D, align, rng)
    # Edge decoherence: reduce cross-branch alignment as tau grows
    h = min(1.0, tau/2.0)
    nrg = np.random.default_rng(rng.integers(10**9))
    W56 = (1-h)*W56 + h*rand_matrix(D, 0.9, nrg)
    W35 = (1-h)*W35 + h*rand_matrix(D, 0.9, nrg)

    # Two parallel subpaths (parts), macro map is their sum
    Jb1 = W46 @ W34 @ W13
    Jb2 = W56 @ W35 @ W23
    JM  = Jb1 + Jb2

    EI_macro = ei_proxy(JM, alpha)
    EI_parts = ei_proxy(Jb1, alpha) + ei_proxy(Jb2, alpha)
    dEI_g = max(0.0, EI_macro - EI_parts) / (EPS + EI_macro)

    # Activations for C_sh with node noise
    a1 = rng.normal(size=D); a2 = rng.normal(size=D)
    a3 = W13@a1 + W23@a2; a4 = W34@a3; a5 = W35@a3; a6 = W46@a4 + W56@a5
    a1o = a1 + tau*rng.normal(size=D); a2o = a2 + tau*rng.normal(size=D)
    a3o = a3 + tau*rng.normal(size=D); a4o = a4 + tau*rng.normal(size=D)
    a5o = a5 + tau*rng.normal(size=D); a6o = a6 + tau*rng.normal(size=D)

    edges = [(a1o,a3o,W13),(a2o,a3o,W23),(a3o,a4o,W34),
             (a3o,a5o,W35),(a4o,a6o,W46),(a5o,a6o,W56)]
    num = den = 0.0
    for au,av,W in edges:
        r = W@au - av; num += r@r; den += au@au + av@av
    Csh = np.sqrt(num) / (EPS + np.sqrt(den))
    EICS = dEI_g / (1.0 + Csh)
    return Csh, dEI_g, EICS

# Sweep tau and average over seeds
C_m, d_m, S_m = [], [], []
C_e, d_e, S_e = [], [], []
for tau in TAUS:
    Cs, ds, Ss = [], [], []
    for k in range(N_SEEDS):
        rng = np.random.default_rng(1000 + k)
        Csh, dEIg, EICS = metrics_at_tau(tau, rng)
        Cs.append(Csh); ds.append(dEIg); Ss.append(EICS)
    Cs, ds, Ss = map(np.array, (Cs,ds,Ss))
    C_m.append(Cs.mean()); d_m.append(ds.mean()); S_m.append(Ss.mean())
    C_e.append(Cs.std(ddof=1)/np.sqrt(N_SEEDS))
    d_e.append(ds.std(ddof=1)/np.sqrt(N_SEEDS))
    S_e.append(Ss.std(ddof=1)/np.sqrt(N_SEEDS))

# Plot (one panel; PDF for LNCS)
TAUS = np.array(TAUS); C_m=np.array(C_m); d_m=np.array(d_m); S_m=np.array(S_m)
C_e=np.array(C_e); d_e=np.array(d_e); S_e=np.array(S_e)
plt.figure(figsize=(6.2,4.2))
plt.plot(TAUS, S_m, '-o', label='EICS')
invC = 1.0/(1.0 + C_m)
plt.plot(TAUS, invC, '--', label=r'$1/(1+C_{\mathrm{sh}})$')
plt.plot(TAUS, d_m, ':', label=r'$\widetilde{\Delta \mathrm{EI}}_G$')
plt.xlabel(r'Noise scale $\tau$'); plt.ylabel('Score (unitless)')
plt.title('Toy sanity-check: effect of noise on EICS and components')
plt.grid(True, linewidth=0.5, alpha=0.5); plt.legend(frameon=False)
plt.tight_layout(); plt.savefig('fig_toy_noise_curve.pdf', bbox_inches='tight')
\end{lstlisting}

\bibliographystyle{splncs04}
\bibliography{eics_refs}

\end{document}